\documentclass[conference]{IEEEtran}
\usepackage{cite}
\usepackage{amsmath,amssymb}
\usepackage{graphicx,booktabs,array}
\usepackage{custom}
\usepackage{cleveref}

\newauthornote{\wei}{Wei}{blue}
\newauthornote{\chang}{Chang}{teal}
\newauthornote{\hy}{Hyerin}{violet} 
\newauthornote{\joe}{Joerg}{orange}

\newcommand{\ptciou}{PTC-IoU}
\newcommand{\Det}{\mathrm{Det}}
\newcommand{\Ass}{\mathrm{Ass}}
\newcommand{\Loc}{\mathrm{Loc}}
\newcommand{\IoU}{\operatorname{IoU}}
\newcommand{\hw}{Jetson Orin Nano Super}
\begin{document}
\raggedbottom
\title{PTC-IoU: CPU-Friendly Detection Quality Estimation for Edge Video Analytics}
\author{\IEEEauthorblockN{Wei Geng, Hyerin Kim, Chang Lu, Zhenghao Lu, Joerg Ott}
\IEEEauthorblockA{Technical University of Munich\\
\{firstname.lastname\}@tum.de}}
\maketitle
\begin{abstract}

An edge video analytics system must estimate the quality of local detections since there are no ground-truth labels available during runtime. Detectors usually provide confidence scores for their results, but these scores can not reveal every missed object or misplaced bounding box. So a quality estimator is necessary to provide a decision signal indicating which frame has potential detection failures that need to be escalated to a more powerful model in the cloud.
Neural network based estimators add workload to the GPU already running detection, while CPU capacity may remain available. We present Predictive Temporal-Consistency IoU (PTC-IoU), which moves this workload to CPU by compacting detection boxes histories to support deterministic result scoring.
Comparing to neural counterparts, experiments on representative edge hardware, Nvidia Jetson Orin Nano, show that our method reduced per-frame estimator latency from 10.3\,ms for neural counterparts to 2.2\,ms (4.7$\times$ faster), increased pipelined throughput by 13.6\%, and cuts estimator energy consumption by up to 95\%, while maintaining competitive final detection quality verified on multiple datasets.
\end{abstract}
\begin{IEEEkeywords}
Edge Computing, 
Video Analytics, 
Selective Offloading, 
ML-Systems Co-design, 
PTC-IoU
\end{IEEEkeywords}
\section{Introduction}

\begin{figure}[t]
    \centering
    \includegraphics[width=\columnwidth]{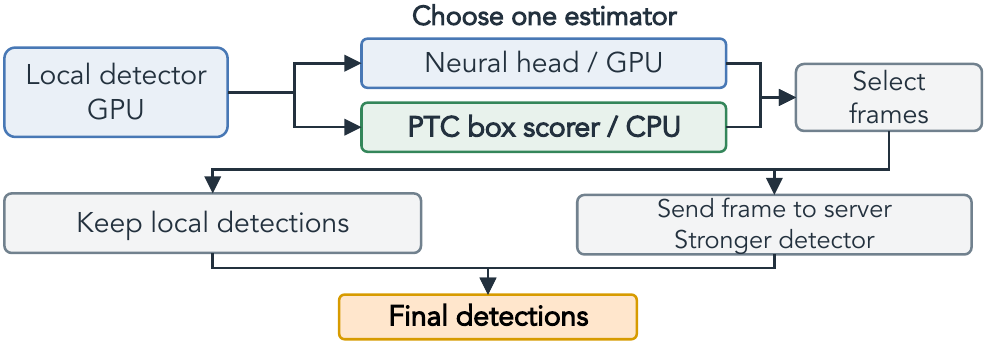}
    \caption{Selective offloading between edge and cloud.}
    \label{fig:selective-offloading}
\end{figure}

A roadside camera paired with a cloud server needs to detect vehicles continuously under limited compute and communication capacity. \Cref{fig:selective-offloading} demonstrates that a lightweight detector on edge devices can process each frame locally, while a stronger model in the cloud can assist with only the difficult images. The device therefore faces a second task alongside detection: deciding which results to keep or send to the cloud for stronger inference. This is called \emph{selective offloading}.

Selective offloading decision requires information that the detector does not directly provide. A detector returns object locations as bounding boxes, together with classes and confidence scores. 
Though a car may be confidently detected, this does not indicate whether another car was missed, or whether the detected box is accurately placed~\cite{jiang2018acquisition}.
During runtime, there are no ground-truth labels against which to check the result. 
In \emph{selective offloading}, \emph{quality estimators} provide a decision signal for choosing only suspicious frames for cloud offloading before spending the communication and server-compute budget, to maximize the final detection quality while lowering the offload cost.

\begin{figure}[th]
 \centering
 \includegraphics[width=\columnwidth]{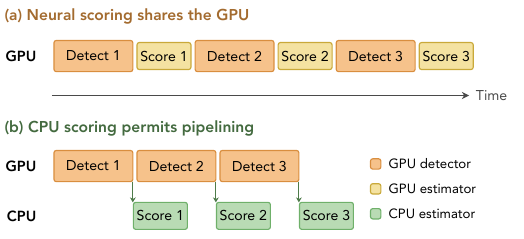}
 \caption{Quality-estimator placement. (a) Neural scoring shares the GPU with detection. (b) CPU scoring can overlap detection of the next frame. Schedules are conceptual and do not represent measured execution times.}
 \label{fig:placement}
 \vspace{-1em}
\end{figure}

Producing this decision signal also consumes edge resources. 
A dedicated neural network head can reuse the detector's features~\cite{jiang2018acquisition,li2020generalized}, making GPU execution alongside detection a natural choice.
As illustrated in \Cref{fig:placement}(a), neural scoring adds a stage to the GPU already running detection.
With detection and GFL quality estimation on the GPU, \Cref{fig:placement-utilization} shows lower average CPU utilization alongside a more heavily used GPU. This opportunity extends beyond Jetson: CoDL improves on GPU-only inference by sharing work with the CPU across Snapdragon and Kirin mobile SoCs~\cite{jia2022codl}.
This suggests a design opportunity and challenge: 
\Cref{fig:placement}(b) proposes to move the quality estimator to the CPU, while keeping the GPU focused on detection. However, existing neural heads require matrix multiplications that are inefficient on the CPU.
Constructing the estimation operations that the CPU can process efficiently is therefore necessary.

We introduce \ptciou{} (Predictive Temporal-Consistency IoU, abbreviated as PTC), which jointly considers the quality signal and its execution features. For example, vehicles generally persist across adjacent video frames, with short-term continuity in position and size. 
PTC defines three cues (detection persistence, association clarity, and localization consistency) using the detector's existing bounding boxes and confidence scores. It replaces image-feature processing and neural scoring with compact box histories and arithmetic on box overlaps and motion.

Our primary contribution is a lightweight detection quality estimation function with CPU-friendly execution: enabling efficient edge deployment while maintaining competitive accuracy. Experimental results show that our estimator achieves competitive quality while cutting per-frame processing latency and energy consumption on an NVIDIA \hw{} testbed. Code and experiments are publicly available\footnote{\faGithub\enspace Source code: \url{https://github.com/cm-edge/byd-acc}.}.


\section{Related Work}
\label{sec:related}
\textbf{Quality signals for selective inference.} SelectiveNet jointly learns prediction and rejection to balance error and coverage~\cite{geifman2019selectivenet}. For object detection, IoU-Net estimates localization confidence~\cite{jiang2018acquisition}, while GFL combines classification and localization quality~\cite{li2020generalized}. PTC instead derives a frame-level signal from existing box histories, avoiding additional neural scoring on the GPU.

\textbf{Temporal cues from detections.} Ramanagopal et al. identify potential missed detections through temporal and stereo inconsistencies, validated by a learned classifier without ground-truth labels at inference~\cite{ramanagopal2018failing}. The IoU tracker associates detections using box geometry without new image features~\cite{bochinski2017high}. HOTA distinguishes detection, association, and localization in ground-truth-based tracker evaluation~\cite{luiten2021hota}. PTC uses detection persistence, association clarity, and localization consistency to produce a deterministic CPU score for frame selection.

\textbf{Reducing the cost of cooperative inference.} NoScope combines temporal difference detectors and specialized cascades for object-presence queries over fixed-camera video~\cite{kang2017noscope}. Cao et al. use a difficult-case discriminator to divide images between local and remote detectors~\cite{cao2023edge}. Qiu et al. estimate offloading quality gains from local detections under an upload budget~\cite{qiu2024optimizing}. Budget-dependent estimator placement can skip local inference for offloaded frames~\cite{geng2026budget}. We focus on the local cost of producing the selection signal, assessing latency, throughput, and energy alongside final detection average precision (AP) at different offload budgets.

\begin{figure}[t]
 \centering
 \includegraphics[width=\columnwidth]{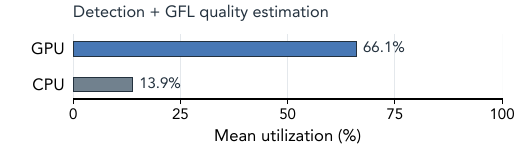}
 \caption{Mean processor utilization with detection and GFL quality estimation on the GPU, measured in the energy run under the bursty workload. CPU utilization is averaged across six cores.}
 \label{fig:placement-utilization}
 \vspace{-1em}
\end{figure}

\textbf{Resource-aware execution.} CoDL partitions neural operators between CPU and GPU~\cite{jia2022codl}, while Band schedules DNN subgraphs across heterogeneous mobile processors~\cite{jeong2022band}. Both optimize a given neural workload. With detection and GFL sharing the GPU (\Cref{fig:placement-utilization}), GPU utilization is 66.1\%, while CPU utilization averages 13.9\% across cores. This imbalance motivates PTC's joint design of the quality representation and scoring operations: arithmetic on compact box histories makes scoring inexpensive on the CPU.

\section{PTC-IoU: CPU-Friendly Temporal Estimation}
\label{sec:method}

\begin{figure*}[t]
 \centering
 \includegraphics[width=\textwidth]{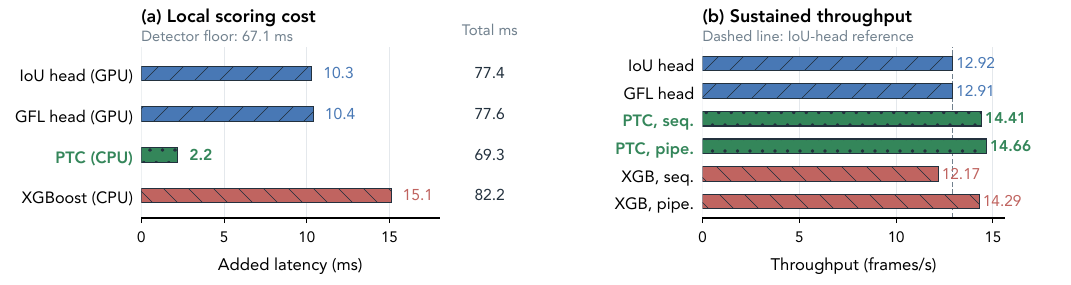}
 \caption{Local processing performance on \hw{} under the bursty workload (\Cref{sec:setup}). (a) Added mean frame time relative to paired detector-only runs, with total latency alongside. (b) Sequential and pipelined sustained throughput.}
 \label{fig:performance}
 \vspace{-1em}
\end{figure*}

PTC estimates frame quality through detection persistence~\Cref{sec:method:pers}, association clarity~\Cref{sec:method:asso}, and localization consistency~\Cref{sec:method:loc}, using the detector's existing boxes, classes, and confidences.
\subsection{Detection: do confident detections persist?}
\label{sec:method:pers}

Detection persistence measures how much detection confidence is supported in both adjacent frames. PTC repeatedly pairs unused same-class boxes with the largest intersection over union (IoU), stopping below 0.5, an empirical threshold selected from $\{0.3,0.5,0.7\}$. Each pair $(i,j)\in\mathcal{M}_t$ links previous-frame box $b_i^{t-1}$ to current-frame box $b_j^t$, with confidences $s_i^{t-1}$ and $s_j^t$. Each box appears in at most one pair.

Let $S_t=\sum_j s_j^t$ be the total confidence in frame $t$, and $C_t=\sum_{(i,j)\in\mathcal{M}_t}\min(s_i^{t-1},s_j^t)$ the confidence shared across matched pairs. Detection persistence is
\begin{equation}
 \Det_t=\frac{C_t}{S_{t-1}+S_t-C_t}.
 \label{eq:det}
\end{equation}
\textbf{Intuition.} The score is a confidence-weighted IoU of the detection sets. A matched pair shares only the smaller of its two confidences. The denominator counts total confidence across both frames with this shared part counted once. For example, suppose two cars remain in view and have detection confidences 0.9 and 0.8 in frame $t-1$. If only the first car is detected and matched in frame $t$ with confidence 0.7, then $S_{t-1}=1.7$ and $S_t=C_t=0.7$, giving $\Det_t=0.7/(1.7+0.7-0.7)\approx0.41$. The score therefore depends on both which detections persist and their confidences. Unmatched detections contribute only to the denominator, while unequal confidences within a pair also lower the score.

\subsection{Association: are correspondences unambiguous?}
\label{sec:method:asso}
Association clarity measures whether a current detection has one distinct counterpart in the previous frame. For current box $b_j^t$, let $u_1(j)$ and $u_2(j)$ be the largest and second-largest IoUs with previous same-class boxes, computed before one-to-one pairing. PTC defines the per-box clarity $a_j$ and frame score $\Ass_t$ as
\begin{equation}
 a_j=\frac{u_1(j)-u_2(j)}{u_1(j)+\epsilon},
 \qquad
 \Ass_t=\frac{\sum_j s_j^t a_j}{S_t}.
 \label{eq:ass}
\end{equation}
\textbf{Intuition.} The score measures the gap between the two best candidates relative to the best overlap. Suppose a car's current box overlaps two previous car boxes with IoUs 0.8 and 0.2. Then $a_j\approx(0.8-0.2)/0.8=0.75$, indicating a distinct best candidate. If the competing overlap rises to 0.75, $a_j\approx(0.8-0.75)/0.8\approx0.06$, indicating ambiguity despite the same best overlap. $\Ass_t$ averages these scores across current detections, weighted by their confidences $s_j^t$. Missing candidates have zero overlap, and the small constant $\epsilon$ prevents division by zero.

\subsection{Localization: do boxes follow plausible motion?}
\label{sec:method:loc}
Localization consistency compares the current detector output $b_j^t$ with a motion estimate $\widehat b_i^t$ derived from past detections. For each pair $(i,j)\in\mathcal{M}_t$, PTC independently finds the box in frame $t-2$ with greatest positive overlap with $b_i^{t-1}$, regardless of class. It repeats their center displacement and width/height ratios to estimate the box at $t$.

\begin{figure}[!ht]
 \centering
 \includegraphics[width=\columnwidth]{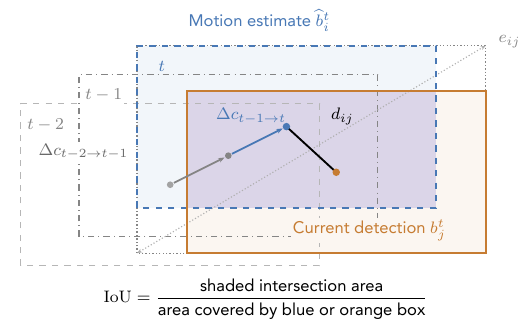}
 \caption{Gray boxes are past detections. Repeating $\Delta c_{t-2\to t-1}$ as $\Delta c_{t-1\to t}$ gives the blue motion estimate, shown with unchanged size. Only this estimate and the orange current detection enter the overlap and center-distance comparison. PTC also extrapolates width and height ratios when size changes.}
 \label{fig:localization}
 \vspace{-1em}
\end{figure}

\Cref{fig:localization} illustrates both extrapolation and comparison. Let $d_{ij}^2$ be the squared center distance between the motion estimate and current detection, and $e_{ij}^2$ the squared diagonal of their enclosing rectangle. Then
\begin{equation}
 \begin{aligned}
  \ell_{ij}&=\IoU(\widehat b_i^t,b_j^t)^\alpha
  \exp\!\left(-\frac{(1-\alpha)d_{ij}^2}{e_{ij}^2+\epsilon}\right),\\
  \Loc_t&=\frac{1}{|\mathcal{M}_t|}
  \sum_{(i,j)\in\mathcal{M}_t}\ell_{ij}.
 \end{aligned}
 \label{eq:loc}
\end{equation}

\textbf{Intuition.} Exact agreement with the motion estimate gives $\ell_{ij}=1$. The IoU term measures agreement in position and size, while the exponential term further penalizes center displacement relative to the enclosing rectangle. With our choice $\alpha=0.7$, an IoU of 0.6 and normalized squared center displacement $d_{ij}^2/(e_{ij}^2+\epsilon)=0.03$ give $\ell_{ij}=0.6^{0.7}\exp(-0.009)\approx0.69$. $\Loc_t$ averages this agreement over matched pairs. Without an older overlapping box, $\ell_{ij}$ uses adjacent-frame IoU.

\subsection{Combining the three perspectives}
The frame score combines the three components:
\begin{equation}
 q_t=\Det_t^{\,w_d}\Ass_t^{\,w_a}\Loc_t^{\,w_l}.
 \label{eq:online}
\end{equation}
Nonnegative weights control each component's contribution. A zero weight omits that component. Higher $q_t\in[0,1]$ indicates stronger temporal consistency, though persistent errors can also score highly.

To obtain $q_t$, PTC filters detections by class and confidence (\Cref{sec:setup}), then evaluates and combines the components on the CPU. Detection and association compare adjacent frames, while localization also uses older detections for motion prediction. Each video's first frame initializes the two-frame history, which PTC updates after scoring subsequent frames.

All components equal one when both adjacent frames have no retained detections, and zero when only one does. Otherwise, no accepted pair gives $\Det_t=\Loc_t=0$.

\section{Evaluation}
\label{sec:evaluation}
\begin{figure*}[t]
 \centering
 \includegraphics[width=\textwidth]{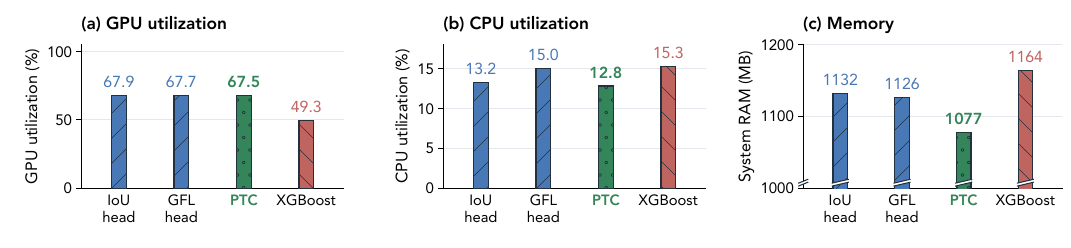}
 \caption{System utilization under the bursty workload and memory footprint. (a) GPU load sampled every 40\,ms. (b) CPU load averaged across six cores. (c) Total system RAM, combining the detector-only baseline and separately measured estimator increments. The broken RAM axis omits 0--1000\,MB. Settings appear in \Cref{tab:setup}.}
 \label{fig:resources}
 \vspace{-1em}
\end{figure*}

\subsection{Experimental setup and comparison conditions}
\label{sec:setup}
\begin{table}[t]
 \centering
 \caption{Hardware benchmarks and offline quality evaluation.}
 \label{tab:setup}
 \small
 \setlength{\tabcolsep}{3pt}
 \renewcommand{\arraystretch}{1.04}
 \begin{tabular}{@{}>{\raggedright\arraybackslash}p{0.24\columnwidth}>{\raggedright\arraybackslash}p{0.72\columnwidth}@{}}
 \toprule
 Item & Setting\\
 \midrule
 Platform & Orin Nano Super, six-core Cortex-A78AE CPU, Ampere GPU. L4T R36.5, MAXN\_SUPER, stock DVFS, Docker/PyTorch 2.2.0\\
 Detector pair & COCO-pretrained Faster R-CNN: MobileNetV3-Large FPN (local), ResNet-50 FPNv2 (stronger)\\
 Local resolution & System tests: FPN-320, min/max side 320/640 pixels. Quality replay: 800/1333 pixels\\
 Placement & \textbf{CPU}: PTC and XGBoost. \textbf{GPU}: IoU and GFL heads\\
 System input & MOT17-02-FRCNN, batch 1, person boxes, confidence 0.1 for every estimator\\
 Timing & Bursty workload, eight rounds, each estimator condition paired with the adjacent detector-only runs. Mean frame time derived from throughput\\
 Utilization & Bursty workload, 20 warm-up + 100 measured frames, six rounds in alternating condition order\\
 RAM & Detector baseline and estimator increments measured separately, medians of five runs\\
 Energy & Single stream. Bursty workload: five rounds. Fixed-rate input (4 frames/s): six rounds\\
 Quality splits & Train/test sequences: UA-DETRAC 12/40, UAVDT 12/8, MOT17 4/3 (train 04/05/09/10, test 02/11/13)\\
 Quality metric & AP at IoU 0.7. Mean frame AP for vehicles, pooled AP for MOT17 pedestrians\\
 \bottomrule
 \end{tabular}
 \vspace{-1em}
\end{table}

We assess the resource and energy cost of producing a quality signal on \hw{}~\cite{nvidia2023orin} and its usefulness for frame selection in offline replay. \Cref{tab:setup} summarizes the two evaluation settings.

\textbf{Baselines.} The local detector combines Faster R-CNN~\cite{ren2015faster} and MobileNetV3-Large~\cite{howard2019searching}. The GPU alternatives are an IoU-regression head inspired by IoU-Net~\cite{jiang2018acquisition} and a GFL-based head trained with quality focal loss~\cite{li2020generalized}. Both reuse backbone features, pooling regions to $7\times7\times256$ before two 1024-unit hidden layers. An XGBoost CPU predictor~\cite{chen2016xgboost}, including temporal feature extraction, tests whether CPU placement alone provides inexpensive scoring. Each system comparison includes the complete scoring path.

\textbf{Matched system workload.} All system benchmarks use the FPN-320 detector on MOT17-02-FRCNN, with the same person boxes at confidence 0.1 supplied to every estimator. PTC uses the vectorized product $q_t=\Det_t\Ass_t\Loc_t$ without temporal averaging. Timing, utilization, RAM, and energy use separate benchmark runs in case of interference.
Totals and paired increments are rounded separately. The quality replay retains the original full-resolution detector pair and uses the replay settings described below.

Timing and utilization use a \emph{bursty workload} with no imposed inter-frame delay to measure sustained processing capacity. Energy measurements further use \emph{fixed-rate input} at 4 frames/s (250\,ms intervals) to compare methods at typical steady workload demand.

\textbf{Video workloads and replay.} UA-DETRAC covers roadside vehicles~\cite{wen2020ua}, UAVDT adds aerial vehicle views and camera motion~\cite{du2018unmanned}, and MOT17 covers crowded pedestrian scenes~\cite{dendorfer2021motchallenge}. Neural-head training includes MOT17-specific training. Replay ranks frames within each evaluation partition and substitutes cached stronger-detector results for the selected fraction. AP uses IoU 0.7, averaged per frame for the vehicle datasets and pooled across detections for MOT17. PTC retains boxes at confidence 0.5 on UA-DETRAC and 0.1 on UAVDT and MOT17 according to their different object sizes.

\begin{figure}[t]
 \centering
 \includegraphics[width=\columnwidth]{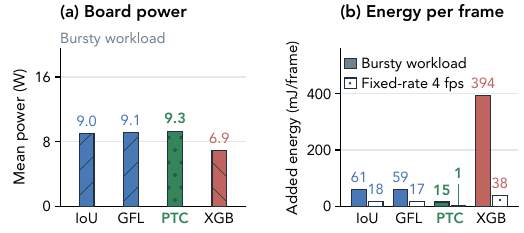}
 \caption{Single-stream energy benchmarks. (a) Board power under the bursty workload. (b) Estimator-added energy per frame under the bursty workload and fixed-rate input (4 frames/s). The fixed-rate PTC increment is estimated by repeated scoring. Detector-only baselines are 632\,mJ/frame (bursty workload) and 1454\,mJ/frame (4 frames/s).}
 \label{fig:energy}
 \vspace{-1em}
\end{figure}

\subsection{Local processing latency and throughput}
We measured per frame estimation latency by calculating the difference in mean frame time between each estimator condition and its paired detector-only runs.
\Cref{fig:performance}(a) shows that PTC needs 2.2\,ms per frame, compared with 10.3--10.4\,ms for the GPU heads, accelerating about $4.7\times$ faster.
The reported total is 69.3\,ms for PTC versus 77.4--77.6\,ms for the GPU heads, a 10.5--10.7\% reduction, or about 8\,ms saved per frame.
\Cref{fig:performance}(b) shows sustained throughput under the bursty workload. The comparison includes pipelined and non-pipelined versions of PTC and XGBoost. The GPU heads reach 12.91--12.92 frames/s, while PTC reaches 14.41 frames/s, an 11.5--11.6\% increase.

\textbf{Computation and placement matter together.} 
Due to neural-head overhead and GPU executing features, baselines like IoU-head and GFL-head incure over 10\,ms of added latency. PTC's compact representation suits CPU execution which has better speed for arithmetic and sequential operations (cpu has higher clock speed and lower latency for those operations). 
XGBoost is also happening on CPU but has a more complex model and feature extraction because its learning based approach requires more computation.

As a consequence, PTC's lightweight scoring implementation achieves higher throughput compared to the GPU heads. Except PTC's improvement in throughput over baselines, there are two worth noting points: 
a) Placement is not the only factor: XGBoost runs on CPU but has a significant overhead, which limits its throughput at 12.17 frames/s.
 PTC's simplified computation contributes to its lower latency and higher throughput, even though it also runs on CPU.
b) Pipelining: PTC's pipelined and non-pipelined throughput is very close since its scoring overhead is already low, the benefit of pipelining is limited. As contrasted, XGBoost has a significant overhead on CPU, so pipelining remove the CPU bottleneck and allows the GPU to process frames while CPU is scoring, which increases throughput. This also verifies XGBoost's CPU bottleneck.
The detector-only baseline is 67.1\,ms per frame, which means if we select a more lightweight detector, the improvement in both latency and throughput from PTC over the GPU heads can be even more significant. However, XGBoost's will not benefit too much since its CPU bottleneck will account for a larger fraction of the total frame time.

\begin{figure*}[t]
 \centering
 \includegraphics[width=\textwidth]{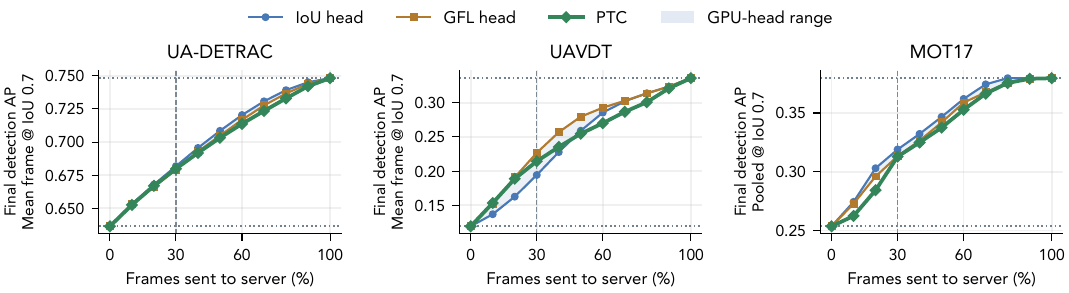}
 \caption{Final detection quality at equal server budgets. Each point combines stronger detections for selected frames with local detections for the rest. Compare curves vertically at the same fraction, where higher AP means better final results. Shading spans the GPU-head curves, and dashed lines mark \Cref{tab:quality}'s 30\% comparison. Dotted lines mark local-only and stronger-only results. Scoring settings appear in \Cref{sec:setup}. AP uses IoU 0.7.}
 \label{fig:routing}
 \vspace{-1em}
\end{figure*}

\subsection{Processor utilization and memory footprint}
\Cref{fig:resources} reports whole-system load under the bursty workload used to measure sustained throughput at its full speed.
GPU utilization is sampled every 40\,ms, and CPU utilization is averaged across six cores. 
PTC keeps average GPU utilization close to the neural baselines (67.5 vs 67.7--67.9\%), while CPU utilization is surprisingly lowest among all estimators, counter-intuitive to what we might expect from a CPU-based estimator.

Compared with the GPU heads, it sustains higher throughput with similar GPU utilization and slightly lower CPU load, cross-vrifying PTC's lower per-frame processing latency.
PTC's low CPU load is consistent with the work required by each scoring path. Although the neural heads execute tensor arithmetic on the GPU, the CPU still sets up and dispatches RoI pooling and prediction operators and aggregates the returned scores. PTC computes directly from compact box histories, avoiding these additional GPU dispatches and feature-tensor handling. Process-time measurements support this interpretation: PTC adds 1.7\,core-ms of CPU work per frame, compared with 5.2--5.3\,core-ms for the GPU heads, summing CPU time across process threads. This measures the complete scoring path rather than attributing its cost to individual host operations.

While XGBoost has lower throughput, it instead uses most CPU among all estimators. Lower GPU utilization reveals that its low efficiency of GPU utilization is due to the CPU bottleneck blocking GPU running on full speed.
Regarding the memory footprint shown in \Cref{fig:resources}(c), PTC's compact computing representation and small box history result in a smaller memory footprint compared to the GPU heads and XGBoost which require more memory for their more complex features and inputs.



\subsection{Energy cost per completed frame}
\Cref{fig:energy} uses VDD\_IN board power to compare energy cost under the bursty workload and fixed-rate input (4 frames/s). The former tests operation at sustained processing capacity while the latter compares methods under the same steady input demand.
As shown in \Cref{fig:energy}(a), under the bursty workload, GPU heads and PTC have similar power draw while XGBoost draws less power because of its sustained throughput has been affected by its CPU bottleneck.
We further calculated the per-frame estimation energy consumption in \Cref{fig:energy}(b),
besides about $4\times$ lower latency under the bursty workload, PTC surprisingly cuts up to 95\% energy consumption per completed frame compared to the GPU heads.

We attribute this to PTC's compact representation and lightweight scoring computation, which doesn't overdraw the GPU and CPU, allocating the system at a more energy-efficient operating frequency. Another possible reason is that PTC's higher throughput allows the system to process more frames in the same amount of time, reducing the board's overall working time for the same workload. Since the power consumption of the other components on the board is relatively fixed, this lowers the marginal cost. 

\subsection{Competitive final detection quality}
\label{sec:evaluation:quality}
\begin{table}[t]
 \centering
 \caption{Final detection AP at a 30\% offload budget.}
 \label{tab:quality}
 \small
 \setlength{\tabcolsep}{3pt}
 \renewcommand{\arraystretch}{1.03}
 \newcommand{\qualitydelta}[1]{%
  \begingroup
  \def\qualitydeltacolor{black!55}%
  \ifdim #1pt>0pt\relax
   \def\qualitydeltacolor{green!45!black}%
  \else\ifdim #1pt<0pt\relax
   \def\qualitydeltacolor{red!70!black}%
  \fi\fi
  \,{\scriptsize\textcolor{\qualitydeltacolor}{$(#1)$}}%
  \endgroup
 }
 \begin{tabular}{@{}lrrr@{}}
 \toprule
 Method & UA-DETRAC & UAVDT & MOT17\\
 \midrule
 Local ($\rho=0$) & 0.636\qualitydelta{-0.034} & 0.119\qualitydelta{-0.065} & 0.254\qualitydelta{-0.036}\\
 \midrule
 Random & 0.670\qualitydelta{0.000} & 0.184\qualitydelta{0.000} & 0.290\qualitydelta{0.000}\\
 IoU head & \textbf{0.682}\qualitydelta{+0.012} & 0.194\qualitydelta{+0.010} & 0.319\qualitydelta{+0.029}\\
 GFL head & 0.679\qualitydelta{+0.009} & 0.227\qualitydelta{+0.043} & 0.313\qualitydelta{+0.023}\\
 XGBoost & \textbf{0.682}\qualitydelta{+0.012} & \textbf{0.231}\qualitydelta{+0.047} & \textbf{0.326}\qualitydelta{+0.036}\\
 PTC (ours) & 0.680\qualitydelta{+0.010} & 0.214\qualitydelta{+0.030} & 0.313\qualitydelta{+0.023}\\
 \midrule
 Stronger ($\rho=1$) & 0.748\qualitydelta{+0.078} & 0.336\qualitydelta{+0.152} & 0.380\qualitydelta{+0.090}\\
 \bottomrule
 \end{tabular}
 \par\smallskip
 \begin{minipage}{\columnwidth}
 \footnotesize
 Entries show AP (0--1, higher is better) and its absolute change from Random:
 \textcolor{green!45!black}{$(+\Delta)$ gain},
 \textcolor{red!70!black}{$(-\Delta)$ loss},
 \textcolor{black!55}{$(0)$ no change}.
 \textbf{Bold} marks the best AP per column at the 30\% budget, including ties.
 Local ($\rho=0$) and Stronger ($\rho=1$) use entirely local and server detections, respectively.
 \end{minipage}
 \vspace{-1em}
\end{table}

\textbf{Useful selection at a low scoring cost.} In \Cref{fig:routing}, we plot the final end-to-end detection quality (AP) after combining local and server results at different offload budgets. 
Higher AP at the same budget means better final detections with the same fraction of frames sent to the server. The endpoints use entirely local or entirely stronger detections. 
We rank the estimator scores and select the top fraction of frames for stronger detections, and then calculate the final AP after combining the stronger detections for selected frames with local detections for the rest. For example, at the 30\% offload budget, \Cref{tab:quality} shows that PTC improves on local-only detection and random selection across all three datasets, while remaining within 0.002--0.013 AP of the better GPU head. These quality results complement the approximately $4.7\times$ scoring speedup measured in the separate FPN-320 hardware benchmarks (\Cref{fig:performance}).

At most fractions, PTC's final AP on UA-DETRAC and MOT17 is close to that of the better GPU head. On UAVDT, PTC is slightly below the better GPU head at higher offload fractions. We attribute this to that UAVDT's aerial views are generally more challenging, and the it's fast moving camera increase the complexity of the temporal context. For example, our design of $Loc_t$ (\Cref{sec:method:loc}) assumes that the motion of the object is relatively smooth, which may not hold in UAVDT. 
The other two datasets are more stable and PTC's temporal context is sufficient to provide a useful quality signal.



\section{Conclusion}
\label{sec:conclusion}
\ptciou{} demonstrates the value of designing a quality signal together with its processor placement. Compact temporal box histories make auxiliary scoring inexpensive on the CPU while still providing useful information for selective offloading. Results show that an efficient CPU estimator instead reduces CPU usage. As a consequence, the energy yields the most significant improvement.

The quality results in \Cref{sec:evaluation:quality} vary across datasets, with UAVDT suggesting sensitivity to rapid camera motion. Future work will extend temporal cues to handle rapid camera motion and persistent detection errors, and adapt offloading decisions to changing network and energy budgets.

\bibliographystyle{IEEEtran}
\bibliography{references}
\end{document}